\documentclass[letterpaper]{article} 
\usepackage{aaai23}  
\usepackage{times}  
\usepackage{helvet}  
\usepackage{courier}  
\usepackage[hyphens]{url}  
\usepackage{graphicx} 
\usepackage{subfig}
\usepackage{booktabs}
\usepackage{natbib}  
\usepackage{caption} 
\usepackage{comment}
\usepackage{amsmath}
\usepackage{tikz}
\usepackage{pgfplots, pgfplotstable}
\usepgfplotslibrary{colormaps}%
\pgfplotsset{compat=1.15}
\pgfplotsset{
    colormap={slategraywhite}{
        rgb255=(112,128,144)
        rgb255=(255,159,101)
    }}
    
\usepackage{algorithm}
\usepackage{algorithmic}

\usepackage{newfloat}
\usepackage{listings}
\DeclareCaptionStyle{ruled}{labelfont=normalfont,labelsep=colon,strut=off} 
\floatstyle{ruled}
\newfloat{listing}{tb}{lst}{}
\floatname{listing}{Listing}
\usepackage{nameref}

\title{Process Knowledge-infused Learning for Clinician-friendly Explanations}
\author{
    Kaushik Roy, \textsuperscript{\rm 1}
    Yuxin Zi, \textsuperscript{\rm 1}
    Manas Gaur, \textsuperscript{\rm 2} 
    Jinendra Malekar, \textsuperscript{\rm 1}
    Qi Zhang, \textsuperscript{\rm 1}
    Vignesh Narayanan, \textsuperscript{\rm 1}
    Amit Sheth, \textsuperscript{\rm 1}
}
\affiliations{
    \textsuperscript{\rm 1} Artificial Intelligence Institute, University of South Carolina\\ Columbia, South Carolina, US\\
    \textsuperscript{\rm 2} University of Maryland, Baltimore County, US\\
    \{kaushikr, yzi\}@email.sc.edu, manas@umbc.edu, jmalekar@mailbox.sc.edu, qz5@cse.sc.edu,\{vignar,amit\}@sc.edu
}

\usepackage{bibentry}

\pgfplotstableread[col sep=comma,]{data.csv}\datatable

\pgfplotstableread[col sep=comma,]{data2.csv}\datatables

\pgfplotstableread[col sep=comma,]{data3.csv}\datatabless

\pgfplotstableread[col sep=comma,]{data4.csv}\datatablesss

\begin{document}

\maketitle

\begin{abstract}
Language models have the potential to assess mental health using social media data. By analyzing online posts and conversations, these models can detect patterns indicating mental health conditions like depression, anxiety, or suicidal thoughts. They examine keywords, language markers, and sentiment to gain insights into an individual's mental well-being. This information is crucial for early detection, intervention, and support, improving mental health care and prevention strategies. However, using language models for mental health assessments from social media has two limitations: (1) They do not compare posts against clinicians' diagnostic processes, and (2) It's challenging to explain language model outputs using concepts that the clinician can understand, i.e., clinician-friendly explanations. In this study, we introduce Process Knowledge-infused Learning (PK-iL), a new learning paradigm that layers clinical process knowledge structures on language model outputs, enabling clinician-friendly explanations of the underlying language model predictions. We rigorously test our methods on existing benchmark datasets, augmented with such clinical process knowledge, and release a new dataset for assessing suicidality. PK-iL performs competitively, achieving a 70\% agreement with users, while other XAI methods only achieve 47\% agreement (average inter-rater agreement of 0.72). Our evaluations demonstrate that PK-iL effectively explains model predictions to clinicians.
\end{abstract}

\section{Introduction}
\begin{figure*}[!ht]
    \centering
    \includegraphics[width=\linewidth,trim=0cm 0cm 1.0cm 0cm,clip]{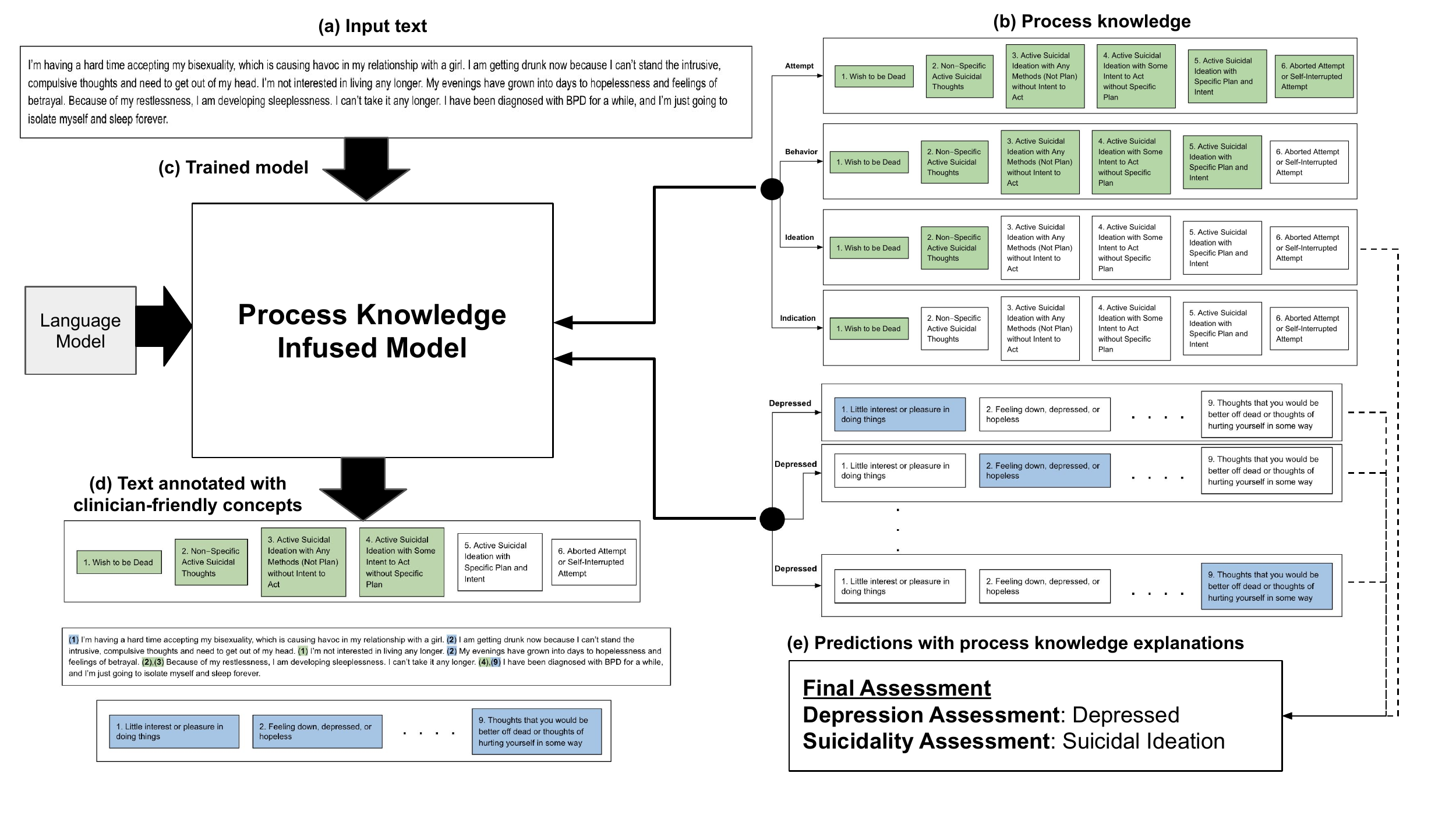}
    \caption{Overview of PKiL inference for an input text. The model uses two arguments, the input text (a) and the process knowledge (b). The process knowledge shows conditions that must be satisfied for a given label. The green part shows process knowledge conditions for suicidality assessment, and the blue part shows the same for depression assessment. For example, for the label \textit{attempt} in the suicidality assessment process knowledge, all conditions 1-6 need to be satisfied. For the label \textit{indication}, only condition 1 needs to be satisfied. The model then annotates text fragments with clinician-friendly concepts from the process knowledge, as shown in (d). The final assessment predictions are obtained through the relevant process knowledge conditions that apply, as shown in (e).}
    \label{fig:overview}
\end{figure*}

A long-standing problem in adopting language models for clinician assistance has been the lack of clinician-friendly explanations for the model's predictions \footnote{\url{https://globelynews.com/world/chatgpt-ai-ethics-healthcare/}}. In practice, a clinical guideline or process is often detailed by which the clinician can assess or label patients. For example, to label patients for degrees of suicidal tendencies in a physical clinical setting, a well-known scale, the Columbia Suicide Severity Rating Scale (CSSRS) \cite{bjureberg2021columbia}, is used to determine the right set of labels. The green part of Figure \ref{fig:overview} (b) shows the CSSRS scale, a \textit{process}, which consists of six conditions whose values determine four assessment outcomes from the set \{\textit{indication}, \textit{ideation}, \textit{behavior}, \textit{attempt}\}.
Similarly, when patients are assessed for depression, clinicians evaluate patient responses against a process or guideline like the Patient Health Questionnaire-9 (PHQ-9) and provide explanations for their assessment using the same. The blue part of Figure \ref{fig:overview} (b) shows the PHQ-9 assessment process. Language models do not explicitly leverage such process knowledge to derive their predictions. Furthermore, language model predictions are typically explained using XAI methods, such as LIME and SHAP, which fits a simpler interpretable surrogate model \cite{adadi2018peeking,ribeiro2016should,lundberg2017unified}. XAI models, however, provide explanations that benefit computer scientists in debugging and improving language models but are of limited utility to the clinician for making decisions. Additionally, it is challenging to approximate very large and complex models, e.g., language models (LMs) using simpler surrogate models \cite{vaswani2017attention}.

We propose a novel learning framework \textit{Process Knowledge infused Learning} (PKiL) that leverages explicit representations of publicly available knowledge of processes and guidelines to augment language models to enable clinician-friendly explanations. Crucially, PKiL incorporates process knowledge structures to provide explanations for model predictions using concepts that are familiar to a clincian. Figure \ref{fig:overview} shows the execution flow of a model trained using PKiL. The PKiL learning framework achieves this through a novel training method with the following salient features - (1) PKiL leverages powerful language models with hundreds of millions of parameters while requiring training of very few additional parameters (equal to the number of process knowledge conditions, e.g., conditions in Figure \ref{fig:overview} (b)) to obtain clinician-friendly explanations, (2) The optimization objective is simple to understand, enabling globally optimal solution discovery through various optimization procedures.

\section{Problem Formulation, Resource Construction, and Process Knowledge infused Learning}
\subsection{Problem Formulation}
Let $X_{\mathcal{D}}$ denote a dataset of input texts and their labels in a domain $\mathcal{D}$. An example of an input post is shown in Figure \ref{fig:overview} (a), and its suicidality assessment label is from the set \{\textit{indication}, \textit{ideation}, \textit{behavior}, \textit{attempt}\} in the domain of mental health. Let $Pk_{\mathcal{D}}$ denote the relevant process knowledge available to us from established literature in domain $\mathcal{D}$. For example, Figure \ref{fig:overview}(b) shows the process of obtaining suicidality assessment labels. Let $\Lambda_{\mathcal{D}}$ be a language model available to us that is fine-tuned on domain specific data (e.g., BERT fine-tuned on mental health posts from social media).  Process Knowledge infused Learning (PKiL) is a training method that makes combined use of $X_{\mathcal{D}}$ and $Pk_{\mathcal{D}}$ to evaluate the conditions in the process knowledge to predict the final label. The evaluated conditions in the process knowledge are familiar to clinicians and therefore enable clinician-friendly explanations for predictions, as shown in Figure \ref{fig:overview}.  

\subsection{Resource Construction - Construction of Process Knowledge Augmented Datasets}
Due to the recent push for explainable and trustworthy AI, recent studies have published new datasets with knowledge of established processes and guidelines used in a particular domain. For example, Gupta et al. constructed the PRIMATE dataset, which includes a series of depression-related posts labeled by human annotators by checking against the PHQ-9 depression assessment process knowledge \cite{gupta2022learning}. Roy et al. construct the ProKnow dataset that consists of similar process knowledge for question generation (e.g., generate questions about symptoms before causes)  while eliciting mental health-related conversation for psychometric test evaluations \cite{roy2023proknow}. We call such datasets process knowledge augmented datasets \cite{sheth2022process}. Gaur et al. used the CSSRS, the suicidality assessment process knowledge, to annotate suicidality labels for a set of Reddit posts extracted from suicidality-related subreddits \cite{gaur2021characterization}. We will call this dataset CSSRS 1.0, an example of $X_{\mathcal{D}}$ in our problem formulation. Even though their work labeled the posts using the process knowledge contained in the CSSRS as annotation guidelines, the exact process knowledge $Pk_{\mathcal{D}}$ used per data point was not stored. Therefore, we first obtain the $Pk_{\mathcal{D}}$ using the following procedure:
\begin{enumerate}
    \item First, we fine-tune the models Word2Vec, SBERT, RoBERTa, T5, ERNIE, and Longformer  on the CSSRS 1.0 dataset, i.e., the $\Lambda_{\mathcal{D}}$ in our formulation \cite{mikolov2013efficient,reimers2019sentence,liu2019roberta,raffel2020exploring,zhang2019ernie,beltagy2020longformer}.
    \item Second, we evaluate each post in the CSSRS 1.0 dataset against the CSSRS $Pk_{\mathcal{D}}$ conditions using cosine similarity between the fine-tuned representations of the posts and the conditions. Condition evaluation returns 1.0 if the condition is satisfied, else 0.0. We set the similarity threshold to 0.5. We do this for all the models and use the max similarity that is greater than the threshold of 0.5.
    \item Next, we obtain a label for each post in $X_{\mathcal{D}}$ from the set \{\textit{indication}, \textit{ideation}, \textit{behavior}, \textit{attempt}\} by comparing the evaluated condition values against the $Pk_{\mathcal{D}}$. For example, if only condition 1, which is \textit{wish to be dead} evaluates to 1.0, the label is \textit{indication} (see Figure \ref{fig:overview} (b)).
    \item Lastly, we provide our labels to three domain experts and task them with either retaining the labels or editing the labels by referring to the CSSRS $Pk_{\mathcal{D}}$ while recording the inter-rater agreement.
\end{enumerate}
The domain experts in the study checked through the labels of 448 Reddit posts in $X_{\mathcal{D}}$. They edited 235/448 posts and provided the relevant process knowledge conditions 1-6, evaluated during the edit. 
A substantial inter-rater agreement of 0.84 was recorded. Crucially, we augment the CSSRS 1.0 to include the specific process knowledge used for the edited label. We call this new dataset CSSRS 2.0. Examples from the dataset can be found at the link in the footnote\footnote{\url{https://anonymous.4open.science/r/MenatalHealthAnoynomous-8CC3/cssrs\%202.0.csv}}. We will use $X^{Pk}_{\mathcal{D}}$ to denote process knowledge augmented datasets. Note that $|X^{Pk}_{\mathcal{D}}| \leq |X_{\mathcal{D}}|$. For example, CSSRS 2.0 has 235 data points, whereas CSSRS 1.0 has 448 data points. Our experiments use CSSRS 2.0 and PRIMATE.
\subsection{Process knowledge infused Learning}
Consider a single condition process knowledge $Pk_{\mathcal{D}}$ to predict a binary label $L$ for an input $x \in X^{Pk}_{\mathcal{D}}$:
\begin{align*}
    if~(C(x)=1), L(x)=1 \\
    else, L(x)=0
\end{align*}
Here $C(x)$ is a condition evaluation function for the input $x$ that evaluates to $1.0$ if the condition is satisfied and $0$ if the condition is not satisfied. $Pk_{\mathcal{D}}$ can be written algebraically as: 
\begin{equation}\label{eqn:algebraic}
\begin{split}
L(x) &= \mathbf{I}(L(x)=1)(C(x)=1) \\
&+ \mathbf{I}(L(x)=0) 
\end{split}
\end{equation}Here $\mathbf{I}(L(x)=l)$ is the indicator function that evaluates to $1$ or $0$, indicating whether the value that the label $L(x)$ takes is equal to $l$. How do we mathematically formulate $C(x)=1$?
We can parameterize $C(x)=1$ as $S(e^{\Lambda_{\mathcal{D}}}_x, e^{\Lambda_{\mathcal{D}}}_C) \geq \theta_C$, where $S$ is a similarity function (e.g., cosine similarity) and $\theta_C$ is the similarity threshold. The $e^{\Lambda_{\mathcal{D}}}_x$ and $e^{\Lambda_{\mathcal{D}}}_C$ are embeddings of the input and condition obtained using a domain-specific fine-tuned language model $\Lambda_{\mathcal{D}}$. Thus, we can write a parameterized approximation to \eqref{eqn:algebraic} as:
\begin{equation}\label{eqn:sim}
\begin{split}
\hat{L}(x,\theta_C) &= \mathbf{I}(L(x)=1)S(e^{\Lambda_{\mathcal{D}}}_x, e^{\Lambda_{\mathcal{D}}}_C) \geq \theta_C) \\
&+ \mathbf{I}(L(x)=0)
\end{split}
\end{equation}
Now we consider a slightly more complex process knowledge $Pk_{\mathcal{D}}$, a multilabel and multi-conditioned process knowledge to predict label $L \in \{1,2,3\}$, given conditions $C1, C2, C3$, for an input $x \in X^{Pk}_{\mathcal{D}}$:
\begin{align*}
    if~(C1(x)=1 \land C2(x) = 1), L(x) = 1 \\
    if~(C1(x)=1 \land C3(x) = 1), L(x) = 2 \\
    else, L(x) = 3
\end{align*}
Similar to \eqref{eqn:algebraic}, we can write this $Pk_{\mathcal{D}}$ algebraically as:
\begin{equation}\label{eqn:algebraic2}
\begin{split}
L(x) &= \mathbf{I}(L(x)=1)(C1(x)=1)(C2(x)=1) \\
& + \mathbf{I}(L(x)=2)(C1(x)=1)(C3(x)=1) \\
& + \mathbf{I}(L(x)=3)
\end{split}
\end{equation}
Following a similar procedure as the one used to derive \eqref{eqn:sim}, we obtain:
\begin{equation}\label{eqn:sim2}
\begin{split}
& \hat{L}(x, \theta_{C1},\theta_{C2}) = \\
& \mathbf{I}(L(x)=1)(S(e^{\Lambda_{\mathcal{D}}}_x, e^{\Lambda_{\mathcal{D}}}_{C1}) \geq \theta_{C1})) (S(e^{\Lambda_{\mathcal{D}}}_x, e^{\Lambda_{\mathcal{D}}}_{C2}) \geq \theta_{C2})) \\
& + \mathbf{I}(L(x)=2)(S(e^{\Lambda_{\mathcal{D}}}_x, e^{\Lambda_{\mathcal{D}}}_{C1}) \geq \theta_{C1})(S(e^{\Lambda_{\mathcal{D}}}_x, e^{\Lambda_{\mathcal{D}}}_{C3}) \geq \theta_{C3})) \\
& + \mathbf{I}(L(x)=3)
\end{split}
\end{equation}
Generally, given multi-condition process knowledge $Pk_{\mathcal{D}}$ for multilabel prediction of the form
\[if~\land_j(C_j(x)=1), L(x) = l\]
we get its algebraic form as
\begin{equation}\label{sim:algebraic3}
    L(x) = \mathbf{I}(L(x)=l)\prod_j(C_j(x)=1)
\end{equation}
Denoting all the parameters as the set $\{\theta_{C_j}\}$ we get the parameterization
\begin{equation}\label{eqn:sim3}
    \hat{L}(x, \{\theta_{C_j}\}) = \mathbf{I}(L(x)=l)\prod_j(S(e^{\Lambda_{\mathcal{D}}}_x, e^{\Lambda_{\mathcal{D}}}_{C_j}) \geq \theta_{C_j})
\end{equation}
For all $x \in X^{Pk}_{\mathcal{D}}$, we get a system of equations like \eqref{eqn:sim3}. 
\subsubsection{Sentiment Analysis}
The conditions in the process knowledge help the model assess problem issues. However, a complete mental health assessment usually also involves the identification of signs of positivity. Therefore for each $\theta_{C_j}$, we also optimize for a $\gamma_{C_j}$ term, where the model predicts positive sentiment in the input if $S(e^{\Lambda_{\mathcal{D}}}_x, e^{\Lambda_{\mathcal{D}}}_{C_j}) \leq \theta_{C_j} + \gamma_{C_j}$.
\subsubsection{Optimization Problem Formulation}
For a process knowledge augmented dataset $X^{Pk}_{\mathcal{D}}$, we know the ground truths $L(x)$ for all $x \in X^{Pk}_{\mathcal{D}}$. We want to solve for the unknown parameters $\theta_{C_j}$ that yields minimum error between the parameterized approximation $L(x,\{\theta_{C_j}\})$ and the ground truth $L(x)$ i.e., \[\sum_{x\in X^{Pk}_{\mathcal{D}}}\mathcal{E}(\hat{L}(x,\{\theta_{C_j}\}),L(x))\]
Here $\mathcal{E}$ denotes the error function. The choice of similarity functions $S$ is a hyperparameter (We explore cosine similarity and normalized Gaussian kernels in our experiments).

\textit{\textbf{Projected Newton's method:}} When one of the $\{\theta_{C_j}\}$ are fixed, setting $\mathcal{E}(\hat{L}(x,\{\theta_{C_j}\}), L(x))$ to be the cross entropy loss reduces to a strongly convex objective that can be solved by \textbf{Newton's method} (with $\varepsilon$ corrections for low determinant Hessians). After each optimization step, we project the $\theta_{C_j}$ to the $[-1,1]$ range.

\textit{\textbf{Grid Search:}} Since the number of parameters to optimize is small (six for CSSRS 2.0 and nine for PRIMATE), we can perform a grid search over a predefined set of grid values to find the values that yield minimum cross-entropy loss. For our choice of $S$, we choose \textbf{cosine similarity and normalized Gaussian kernel}; therefore, grid search candidate values are in the $[-1,1]$ range.

\textit{\textbf{Optimizing for the $\mathbf{\gamma_{C_j}}$:}} To find the optimal $\gamma_{C_j}$, we first predict positive and negative sentiment labels using the \textbf{Stanford CoreNLP} model for all the inputs.  Next, we perform a grid search in the $[-1,1]$ range and set values for the $\gamma_{C_j}$ that results in the maximum agreement between $S(e^{\Lambda_{\mathcal{D}}}_x, e^{\Lambda_{\mathcal{D}}}_{C_j}) \leq \theta_{C_j} + \gamma_{C_j}$ and the Stanford CoreNLP model labels (only the positive labels).

In our experiments, we try both Newton's method and grid search optimization strategies.

\section{Experiments and Results}
We demonstrate the effectiveness of PkiL training using PRIMATE and CSSRS 2.0 combined with several state-of-the-art language models. We also perform experiments with prompting Text-Davinci-003 using the langchain library\footnote{\url{https://langchain.readthedocs.io/en/latest/}}.
\subsection{Process Knowledge Augmented Datasets}
For CSSRS 2.0, the process knowledge is shown in Figure \ref{fig:overview} (b) (the green part). We input this process knowledge in the form\footnote{Examples can be found at the link: \url{https://anonymous.4open.science/r/MenatalHealthAnoynomous-8CC3/cssrs_annotate.txt}}:
\begin{equation*}
\footnotesize
\begin{split}
    & if~((C1(x),C2(x),C3(x),C4(x),C5(x),C6(x)) = 1),\\
    &~L(x) = attempt \\
    & if~((C1(x),C2(x),C3(x),C4(x),C5(x)) = 1),\\
    &~L(x) = behavior \\
    & if~((C1(x),C2(x))=1), L(x) = ideation\\
    & if~(C1(x)=1), L(x)=indication
\end{split}
\end{equation*}
The conditions $C1-C6$ in the CSSRS are:
\begin{equation*}
\footnotesize
\begin{split}
    & C1:~Wish~to~be~dead \\
    & C2:~Non-Specific~Active~Suicidal~Thoughts \\
    & C3:~Active~Suicidal~Ideation~with~Any~Methods\\
    &~(Not~Plan)~without~Intent~to~Act \\
    & C4:~Active~Suicidal~Ideation~with~Some~Intent~to~Act\\ 
    &~without~Specific~Plan\\
    & C5:~Active Suicidal~Ideation~with~Specific~Plan~and~Intent\\
    & C6:~Aborted~Attempt~or~Self-Interrupted~Attempt
\end{split}
\end{equation*}
For PRIMATE, the process knowledge is a set of nine conditions. If any of the conditions evaluate to yes, the depression assessment label is $1$. This is a binary classification task. We input this process knowledge in the form (we collapse conditions $C3-C8$ for brevity):
\begin{equation*}
\footnotesize
\begin{split}
    & if~(C1(x)=1), L(x) = 1\\
    & if~(C2(x)=1), L(x) = 1\\
    & \dots \\
    & if~(C9(x)=1), L(x) = 1\\
    & else, L(x) = 0
\end{split}
\end{equation*}
The conditions $C1-C9$ in the PHQ-9 are:
\begin{equation*}
\footnotesize
\begin{split}
    & C1:~Little~interest~or~pleasure~in~doing~things \\
    & C2:~Feeling down,~depressed,~or~hopeless \\
    & C3:~Trouble~falling~or~staying~asleep,\\
    &~or~sleeping~too~much\\
    & C4:~Feeling~tired~or~having~little~energy\\
    & C5:~Poor~appetite~or~overeating\\
    & C6:~Feeling~bad~about~yourself,\\
    &~or~that~you~are~a~failure,\\
    &~or~have~let~yourself~or~your~family~down\\
    & C7:~Trouble~concentrating~on~things,\\
    &~such~as~reading~the~newspaper~or~watching~television\\
    & C8:~Moving~or~speaking~so~slowly~that\\
    &other~people~could~have~noticed\\
    &~Or~so~fidgety~or~restless~that\\
    &~you~have~been~moving~a~lot~more~than~usual\\
    & C9:~Thoughts~that~you~would~be~better~off~dead\\
    &~or~thoughts~of~hurting~yourself~in~some~way?
\end{split}
\end{equation*}
Examples from the PRIMATE dataset can be found at the link in the footnote \footnote{\url{https://github.com/primate-mh/Primate2022}}.
\subsection{Experimental and Hyperparameter Configurations during Training}
\begin{enumerate}
    \item \textbf{Embedding models for input post and questions: } We use the models Word2Vec, SBERT, RoBERTa, T5, ERNIE, and Longformer fine-tuned on the training data.
    \item \textbf{Similarity function: } We explore the cosine similarity and the normalized Gaussian kernel (input vectors are normalized to be unit length before plugging into the Gaussian kernel). For the normalized Gaussian kernel, we range the scale parameter between $[-1,1]$ in increments of $0.001$.
    \item \textbf{Parameters for grid search: } During grid search optimization, we explore parameters in the $[-1,1]$ range, again in increments of $0.001$.
    \item \textbf{No. of epochs for Newton's optimization method: } We set max epochs of $100$ and experiment with batch sizes of $16$ and $32$ for Newton's method. We train for only $100$ epochs as we have far more equations than unknowns and also perform early stopping if the total parameter differences are less than $0.001$.
\end{enumerate}
\subsection{Text-Davinci-003 Experiment Details}
\begin{figure}[!h]
    \centering
    \includegraphics[width=\linewidth,trim = 0.5cm 9cm 0cm 0cm, clip]{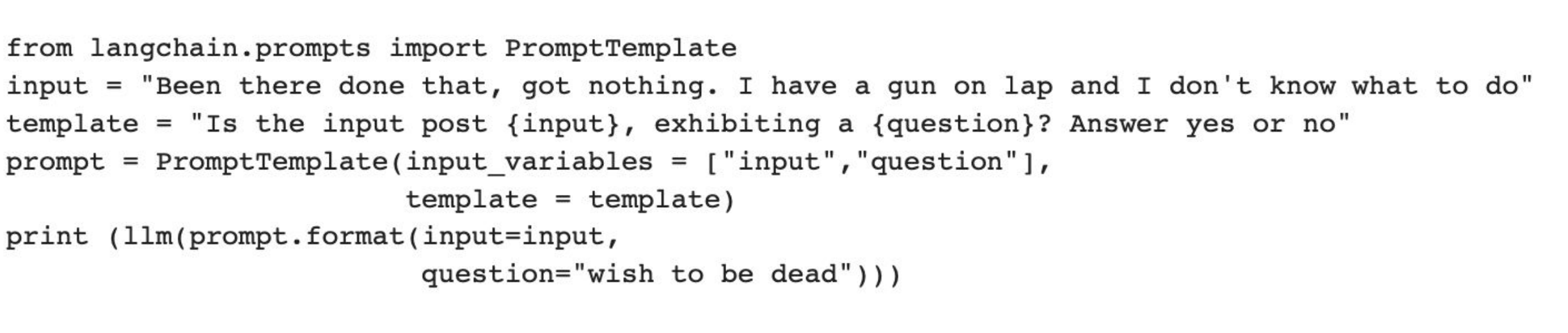}
    \caption{Using the langchain library to prompt Text-Davinci-003 for answers to questions from the process knowledge.}
    \label{fig:gpt}
\end{figure}

We use the langchain library and write a prompt template to obtain answers to the process knowledge questions from Text-Davinci-003. For example, Figure \ref{fig:gpt} shows the prompt template for the first condition $C1:~Wish~to~be~dead$ from the CSSRS process knowledge. For sentiment analysis, we set the \textit{question} variable in Figure \ref{fig:gpt} to \textit{positive sentiment}. We will call this model Text-Davinci-003\textsubscript{PK}.
 Once we evaluate all the conditions, we follow the process knowledge pertaining to the evaluated condition values to determine the label. 
\subsection{Quantitative Results and Discussion}
\begin{figure*}[!htb]
\centering
\includegraphics[width=\linewidth]{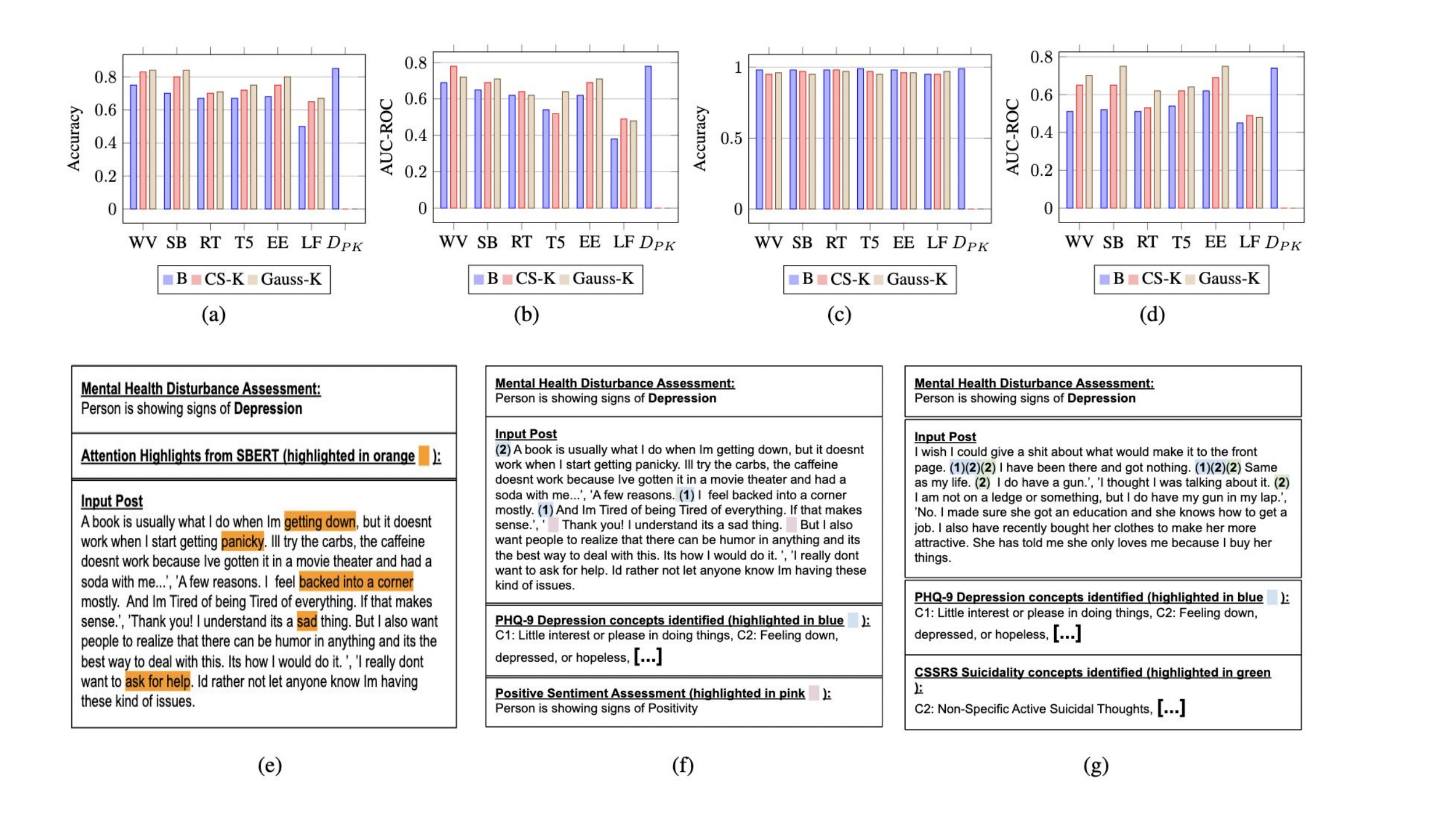}
\caption{(a) and (b) present results for the CSSRS 2.0 dataset, while (c) and (d) show the results for the PRIMATE dataset. The mean accuracy/AUC-ROC of different language models (LMs) - Baseline fine-tuned model (\textbf{B}), PKiL performance with Cosine Similarity Kernel (\textbf{CS-K}), and PKiL performance with normalized Gaussian Kernel similarity (\textbf{Gauss-K}) - are displayed. The prompt-based model Text-Davinci-003\textsubscript{PK} model ($D_{PK}$) doesn't utilize CS-K or Gauss-K, so no associated bar is shown. W2V: Word2Vec, SB: SBERT, RT: RoBERTa, EE: ERNIE, LF: LongFormer. (d) The self-attention-based interpretability visualization for the SBERT baseline model indicates correct predictions and sensible highlights. However, the mapping of these highlights to clinician-friendly concepts used in practice is unclear. Baseline language models consistently struggle to capture negation. (e) The SB model trained with PKiL using the normalized Gaussian kernel provides clinicians with annotated explanations that are more familiar. Additionally, the PKiL parameters enable the analysis of fragments conveying positive sentiment. (f) Explanations from the Text-Davinci-003\textsubscript{PK} model also demonstrate that leveraging process knowledge helps clinicians better understand the annotated explanations, as they are associated with familiar problem concepts.}
\label{fig:results}
\end{figure*}

Figure \ref{fig:results} shows the results of PKiL for various experiment configurations for the CSSRS 2.0 and PRIMATE datasets. The figure also shows results from the Text-Davinci-003\textsubscript{PK} model.

\textit{\textbf{Quantitative Results for CSSRS 2.0: }}
First, excluding the Text-Davinci-003\textsubscript{PK} from the analyses, we observe that SBERT trained using PKiL with a normalized Gaussian kernel performs the best in terms of accuracy, and the Word2Vec model performs the best on AUC-ROC scores for the CSSRS 2.0 dataset. In general, we see that PKiL leads to large boosts in performance of up to 14\% over the baseline. Analysis of The Text-Davinci-003\textsubscript{PK} model performance reveals that it is the best performer among all the models for the CSSRS 2.0 dataset. Our experiments show that large language models can significantly increase suicidality assessment performance when leveraging process knowledge structures and process knowledge-augmented datasets.

\textit{\textbf{Quantitative results for PRIMATE: }}
Again, first excluding the Text-Davinci-003\textsubscript{PK} from the analyses, we observe that RoBERTa trained using PKiL with a cosine similarity function performs the best in terms of accuracy, and SBERT and ERNIE perform the best on AUC-ROC scores for the PRIMATE dataset. In general, we see that PKiL leads to large boosts in performance of up to 23\% over the baseline. Analysis of The Text-Davinci-003\textsubscript{PK} model performance reveals that it is the best performer in terms of accuracy among all the models for the PRIMATE dataset. Our experiments show that large language models can also significantly increase depression assessment performance when leveraging process knowledge structures and process knowledge augmented datasets.
\subsection{Qualitative Results and Discussion}
We evaluate PkiL model outputs qualitatively for the following aspects:
\begin{itemize}
    \item \textbf{Mental health disturbance assessment}: The final label predicted by the model, i.e., the label \textit{depression} for depression assessment), and a label from the set \{\textit{indication}, \textit{ideation}, \textit{behavior}, \textit{attempt}\} for suicidality assessment.
    \item \textbf{PHQ-9 depression concepts identified}: A list of concepts resulting from evaluating conditions C1-C9 using the learned thresholds $\theta_{C_j}$. For the Text-Davinci-003\textsubscript{PK} model, we prompt the model using code as shown in Figure \ref{fig:gpt}.
    \item \textbf{CSSRS suicidality concepts identified}: A list of concepts resulting from evaluating conditions C1-C6 using the learned thresholds $\theta_{C_j}$. Similar to the depression case, for the Text-Davinci-003\textsubscript{PK} model, we prompt the model using code as shown in Figure \ref{fig:gpt}.
    \item \textbf{Positive sentiment assessment}: Using the learned $\theta_j$ and $\gamma_j$ to identify input post fragments that convey positive sentiment.
\end{itemize}

\textit{\textbf{Baseline Model Explanations: }}
We use the bert-viz visualization technique\footnote{\url{https://github.com/jessevig/bertviz}} to interpret the contributions of the different input post fragments to the prediction outcome (the \texttt{CLS} token). Figure \ref{fig:results}(e) shows the output for SBERT. The highlights convey meaningful information from the perspective of depression, which is the correct label. However, it is unclear how the highlights map to clinician-friendly concepts from process knowledge guidelines for depression assessment. A manual post-processing layer for mapping to clinician-friendly concepts is needed in order to verify the prediction.

\textit{\textbf{PKiL Model Explanations: }}
We divide the input post into contiguous fragments of max size $3$ sentences for models and infer the process knowledge condition values using the PKiL trained models and the parameters $\theta_{C_j}$ and $\theta_{\gamma_j}$. We divide for enhanced clinician-friendly explainability as simply annotating the whole posts with concepts still requires additional post-processing by the human to glean out fragments that correspond to problem issues and positive sentiments. Figure \ref{fig:results}(f) shows the output of the SBERT model trained using PKiL with the normalized Gaussian kernel. Figure \ref{fig:results}(g) shows the output of prompting the Text-Davinci-003\textsubscript{PK} as shown in Figure \ref{fig:gpt}. We can readily observe that the explanations are more useful to the clinician as they directly explain the outcome in terms of concepts used in everyday practice. Finally, we provided PKiL explanations to the experts who helped construct the CSSRS 2.0 dataset and asked them to provide the percentage of times they found the explanations beneficial. We also provided baseline explanations for comparison. In order to control for bias, we tell them that humans generate PKiL explanations, and language models generate the baseline explanations. PKiL explanations scored 70\% vs 47\% for the baseline models. We recorded an inter-annotator agreement of 0.72. We analyzed the 30\% that the experts did not find beneficial and observed that models have difficulty distinguishing casual mentions from serious ones. For example, a Reddit user reported wanting to kill themselves out of class boredom before identifying a legitimate clinical issue much further into their post. We leave the investigation of these posts for future work (e.g., by expanding our framework to detect sarcasm). 

\section{Conclusion}
In this study, we develop a novel paradigm PKiL that leverages the combined benefits of explicit process knowledge and high-performance language models to provide predictions and explanations that the end user can readily understand. Our experiments demonstrate the effectiveness of PKiL both quantitatively and qualitatively. Such an improved understanding of language model predictions can inform insights for refining existing process knowledge guidelines (e.g., adaptation to Reddit vocabulary) to facilitate remote monitoring and improved access to healthcare via social media platforms.

\textit{\textbf{Reproducibility: }} We provide the trained model for SBERT with normalized Gaussian kernel similarity, the CSSRS 2.0 dataset, and the CSSRS process knowledge used in our experiments at the link in the footnote\footnote{\url{https://anonymous.4open.science/r/MenatalHealthAnoynomous-8CC3/README.md}}. Additionally, we also provide a Python notebook for users to play with the Text-Davinci-003\textsubscript{PK} model at the link in this footnote\footnote{\url{https://anonymous.4open.science/r/MenatalHealthAnoynomous-8CC3/app.ipynb}}.

\textit{\textbf{Ethics Statement: }} We adhere to anonymity, data privacy, intended use, and practical implication of the AI-based mental health assessment systems. The clinical process knowledge does not contain personally identifiable information. The datasets covered in the survey are publicly available and can be obtained from user-author agreement forms. Figures and examples are abstract and do not represent real-time data sources or any person.
\section{Acknowledgements}
This work was supported in part by the National Science Foundation (NSF) Awards 2133842 “EAGER: Advancing Neuro-symbolic AI with Deep Knowledge- infused Learning,” and was carried out under the advisement of Prof. Amit Sheth \cite{roy2022ksat,roy2022wise,roy2022process,sheth2021knowledge,sheth2022process,sheth2023neurosymbolic}. Any opinions, findings, and conclusions or recommendations expressed in this material are those of the authors and do not necessarily reflect the views of the National Science Foundation.

\bibliography{aaai23}

\begin{thebibliography}{20}
\providecommand{\natexlab}[1]{#1}

\bibitem[{Adadi and Berrada(2018)}]{adadi2018peeking}
Adadi, A.; and Berrada, M. 2018.
\newblock Peeking inside the black-box: a survey on explainable artificial
  intelligence (XAI).
\newblock \emph{IEEE access}, 6: 52138--52160.

\bibitem[{Beltagy, Peters, and Cohan(2020)}]{beltagy2020longformer}
Beltagy, I.; Peters, M.~E.; and Cohan, A. 2020.
\newblock Longformer: The long-document transformer.
\newblock \emph{arXiv preprint arXiv:2004.05150}.

\bibitem[{Bjureberg et~al.(2021)Bjureberg, Dahlin, Carlborg, Edberg, Haglund,
  and Runeson}]{bjureberg2021columbia}
Bjureberg, J.; Dahlin, M.; Carlborg, A.; Edberg, H.; Haglund, A.; and Runeson,
  B. 2021.
\newblock Columbia-Suicide Severity Rating Scale Screen Version: initial
  screening for suicide risk in a psychiatric emergency department.
\newblock \emph{Psychological medicine}, 1--9.

\bibitem[{Gaur et~al.(2021)Gaur, Aribandi, Alambo, Kursuncu, Thirunarayan,
  Beich, Pathak, and Sheth}]{gaur2021characterization}
Gaur, M.; Aribandi, V.; Alambo, A.; Kursuncu, U.; Thirunarayan, K.; Beich, J.;
  Pathak, J.; and Sheth, A. 2021.
\newblock Characterization of time-variant and time-invariant assessment of
  suicidality on Reddit using C-SSRS.
\newblock \emph{PloS one}, 16(5): e0250448.

\bibitem[{Gupta et~al.(2022)Gupta, Agarwal, Gaur, Roy, Narayanan, Kumaraguru,
  and Sheth}]{gupta2022learning}
Gupta, S.; Agarwal, A.; Gaur, M.; Roy, K.; Narayanan, V.; Kumaraguru, P.; and
  Sheth, A. 2022.
\newblock Learning to Automate Follow-up Question Generation using Process
  Knowledge for Depression Triage on Reddit Posts.
\newblock In \emph{Proceedings of the Eighth Workshop on Computational
  Linguistics and Clinical Psychology}, 137--147.

\bibitem[{Liu et~al.(2019)Liu, Ott, Goyal, Du, Joshi, Chen, Levy, Lewis,
  Zettlemoyer, and Stoyanov}]{liu2019roberta}
Liu, Y.; Ott, M.; Goyal, N.; Du, J.; Joshi, M.; Chen, D.; Levy, O.; Lewis, M.;
  Zettlemoyer, L.; and Stoyanov, V. 2019.
\newblock Roberta: A robustly optimized bert pretraining approach.
\newblock \emph{arXiv preprint arXiv:1907.11692}.

\bibitem[{Lundberg and Lee(2017)}]{lundberg2017unified}
Lundberg, S.~M.; and Lee, S.-I. 2017.
\newblock A unified approach to interpreting model predictions.
\newblock \emph{Advances in neural information processing systems}, 30.

\bibitem[{Mikolov et~al.(2013)Mikolov, Chen, Corrado, and
  Dean}]{mikolov2013efficient}
Mikolov, T.; Chen, K.; Corrado, G.; and Dean, J. 2013.
\newblock Efficient estimation of word representations in vector space.
\newblock \emph{arXiv preprint arXiv:1301.3781}.

\bibitem[{Raffel et~al.(2020)Raffel, Shazeer, Roberts, Lee, Narang, Matena,
  Zhou, Li, and Liu}]{raffel2020exploring}
Raffel, C.; Shazeer, N.; Roberts, A.; Lee, K.; Narang, S.; Matena, M.; Zhou,
  Y.; Li, W.; and Liu, P.~J. 2020.
\newblock Exploring the limits of transfer learning with a unified text-to-text
  transformer.
\newblock \emph{The Journal of Machine Learning Research}, 21(1): 5485--5551.

\bibitem[{Reimers and Gurevych(2019)}]{reimers2019sentence}
Reimers, N.; and Gurevych, I. 2019.
\newblock Sentence-BERT: Sentence Embeddings using Siamese BERT-Networks.
\newblock In \emph{Proceedings of the 2019 Conference on Empirical Methods in
  Natural Language Processing and the 9th International Joint Conference on
  Natural Language Processing (EMNLP-IJCNLP)}, 3982--3992.

\bibitem[{Ribeiro, Singh, and Guestrin(2016)}]{ribeiro2016should}
Ribeiro, M.~T.; Singh, S.; and Guestrin, C. 2016.
\newblock " Why should i trust you?" Explaining the predictions of any
  classifier.
\newblock In \emph{Proceedings of the 22nd ACM SIGKDD international conference
  on knowledge discovery and data mining}, 1135--1144.

\bibitem[{Roy et~al.(2023)Roy, Gaur, Soltani, Rawte, Kalyan, and
  Sheth}]{roy2023proknow}
Roy, K.; Gaur, M.; Soltani, M.; Rawte, V.; Kalyan, A.; and Sheth, A. 2023.
\newblock ProKnow: Process knowledge for safety constrained and explainable
  question generation for mental health diagnostic assistance.
\newblock \emph{Frontiers in big Data}, 5: 1056728.

\bibitem[{Roy et~al.(2022{\natexlab{a}})Roy, Gaur, Zhang, and
  Sheth}]{roy2022process}
Roy, K.; Gaur, M.; Zhang, Q.; and Sheth, A. 2022{\natexlab{a}}.
\newblock Process knowledge-infused learning for suicidality assessment on
  social media.
\newblock \emph{arXiv preprint arXiv:2204.12560}.

\bibitem[{Roy et~al.(2022{\natexlab{b}})Roy, Zi, Narayanan, Gaur, Chandrasekar,
  and Sheth}]{roy2022wise}
Roy, K.; Zi, Y.; Narayanan, V.; Gaur, M.; Chandrasekar, S.; and Sheth, A.
  2022{\natexlab{b}}.
\newblock WISE Causal Models: Wisdom Infused Semantics Enhanced Causal Models-A
  Study in Suicidality Diagnosis.

\bibitem[{Roy et~al.(2022{\natexlab{c}})Roy, Zi, Narayanan, Gaur, and
  Sheth}]{roy2022ksat}
Roy, K.; Zi, Y.; Narayanan, V.; Gaur, M.; and Sheth, A. 2022{\natexlab{c}}.
\newblock KSAT: Knowledge-infused Self Attention Transformer--Integrating
  Multiple Domain-Specific Contexts.
\newblock \emph{arXiv preprint arXiv:2210.04307}.

\bibitem[{Sheth et~al.(2021)Sheth, Gaur, Roy, and Faldu}]{sheth2021knowledge}
Sheth, A.; Gaur, M.; Roy, K.; and Faldu, K. 2021.
\newblock Knowledge-intensive language understanding for explainable AI.
\newblock \emph{IEEE Internet Computing}, 25(5): 19--24.

\bibitem[{Sheth et~al.(2022)Sheth, Gaur, Roy, Venkataraman, and
  Khandelwal}]{sheth2022process}
Sheth, A.; Gaur, M.; Roy, K.; Venkataraman, R.; and Khandelwal, V. 2022.
\newblock Process Knowledge-Infused AI: Toward User-Level Explainability,
  Interpretability, and Safety.
\newblock \emph{IEEE Internet Computing}, 26(5): 76--84.

\bibitem[{Sheth, Roy, and Gaur(2023)}]{sheth2023neurosymbolic}
Sheth, A.; Roy, K.; and Gaur, M. 2023.
\newblock Neurosymbolic Artificial Intelligence (Why, What, and How).
\newblock \emph{IEEE Intelligent Systems}, 38(3): 56--62.

\bibitem[{Vaswani et~al.(2017)Vaswani, Shazeer, Parmar, Uszkoreit, Jones,
  Gomez, Kaiser, and Polosukhin}]{vaswani2017attention}
Vaswani, A.; Shazeer, N.; Parmar, N.; Uszkoreit, J.; Jones, L.; Gomez, A.~N.;
  Kaiser, {\L}.; and Polosukhin, I. 2017.
\newblock Attention is all you need.
\newblock \emph{Advances in neural information processing systems}, 30.

\bibitem[{Zhang et~al.(2019)Zhang, Han, Liu, Jiang, Sun, and
  Liu}]{zhang2019ernie}
Zhang, Z.; Han, X.; Liu, Z.; Jiang, X.; Sun, M.; and Liu, Q. 2019.
\newblock ERNIE: Enhanced language representation with informative entities.
\newblock \emph{arXiv preprint arXiv:1905.07129}.

\end{thebibliography}

\end{document}